\documentclass[journal]{IEEEtai}

\usepackage[colorlinks,urlcolor=blue,linkcolor=blue,citecolor=blue]{hyperref}
\usepackage{color,array}
\usepackage{graphicx}
\usepackage{amsmath}
\usepackage{booktabs}
\usepackage{url}
\usepackage{hyperref}
\usepackage{comment}

\begin{document}

\title{Empirical Evaluation of Physical Adversarial Patch Attacks Against Overhead Object Detection Models}

\author{Gavin S. Hartnett, Li Ang Zhang, Caolionn O'Connell, Andrew J. Lohn, Jair Aguirre. 
\thanks{
}
\thanks{
G. S. Hartnett (corresponding author, email: hartnett@rand.org), L. Zhang, C. O'Connell, and J. Aguirre are with the RAND Corporation, Santa Monica, CA 90404 USA.
} 
\thanks{A. Lohn is with Center for Security and Emerging Technology, Georgetown University, Washington, DC, 20005 USA.
}
}

\markboth{Preprint}
{G. S. Hartnett \MakeLowercase{\textit{et al.}}: Empirical Evaluation of Physical Adversarial Patch Attacks Against Overhead Object Detection Models}

\maketitle

\begin{abstract}
Adversarial patches are images designed to fool otherwise well-performing neural network-based computer vision models. Although these attacks were initially conceived of and studied digitally, in that the raw pixel values of the image were perturbed, recent work has demonstrated that these attacks can successfully transfer to the physical world. This can be accomplished by printing out the patch and adding it into scenes of newly captured images or video footage. In this work we further test the efficacy of adversarial patch attacks in the physical world under more challenging conditions. We consider object detection models trained on overhead imagery acquired through aerial or satellite cameras, and we test physical adversarial patches inserted into scenes of a desert environment. Our main finding is that it is far more difficult to successfully implement the adversarial patch attacks under these conditions than in the previously considered conditions. This has important implications for AI safety as the real-world threat posed by adversarial examples may be overstated. 
\end{abstract}

\begin{IEEEImpStatement}
Neural network-based object detection systems have been shown to be vulnerable to a certain kind of digital attack known as adversarial patches. In this work we explore to what extent the systems remain vulnerable when the patches are printed out and applied ``in the real world'', and find that these physical attacks are entirely ineffective.
\end{IEEEImpStatement}


\section{Introduction}

\IEEEPARstart{D}{o} adversarial examples pose a legitimate threat to real-world, safety-critical AI systems? They certainly pose a serious threat in theory; over the past few years adversarial examples have attracted significant attention from the machine learning research community, and thousands of papers have been dedicated both to the problem of devising stronger attacks and defenses against those attacks (see, for example, \cite{yuan2019adversarial}). A common element in these works is the conclusion that machine learning systems are generally vulnerable to adversarial examples. Although bespoke defenses may be devised for any given attack, in general, the defender is at a disadvantage relative to the attacker as newer attack methods can be adopted. A competent and well-resourced attacker could therefore be expected to seriously degrade the performance of machine learning systems through attacks based on adversarial examples, provided they have sufficient access to the target system and/or data. 

The extent of the resources and access available to the attacker is specified as part of the threat model. The threat model determines which attacks are possible in any given scenario. Most often, the threat model is such that the adversarial attacks must be carried out digitally. For example, this scenario could occur if an adversary gained access to a computer network which hosted an image classification model. By intercepting and modifying the digital images which are fed to the model as inputs, the adversary could cause the model to fail or suffer degraded performance. 

However, the above scenario may be a bit unrealistic or impractical. If an adversary gained access to a defender's computer network, there are likely far more destructive actions they could take than add adversarial perturbations to digital images.\footnote{One way to modify this scenario so that adversarial attacks are more relevant would be to stipulate that the images are collected by one service and then submitted to a second service which processes them. In this case, a realistic attack employing adversarial examples might be a man-in-the-middle attack that injected adversarial noise into the images while they were transiting between the two services.}  In contrast, it is much easier to devise realistic scenarios involving “physical” adversarial attacks, which is to say adversarial attacks that have been in some way manifested in the real world. As an example, one such scenario features a self-driving car using neural network-based computer vision models to detect stop signs. By affixing a printed-out version of an adversarial example known as an adversarial patch, an adversary could thwart the stop-sign detector, thus causing a catastrophic incident and demonstrating the severity of the threat posed by adversarial attacks. And, indeed, there have been many successful demonstrations of adversarial attacks in the physical world (as discussed below).

Although adversarial attacks have been shown to transfer to the physical world, there are still many questions about the severity of the threat they pose to safety-critical systems. One key question concerns the robustness of the attacks—how much deviation from the training and testing conditions can the attacks withstand without suffering a serious degradation in their performance? For example, considering adversarial attacks on computer vision systems, a natural question is whether the attacks continue to work under a range of lighting conditions and for a range of camera angles and resolutions. Another question is: how might an adversary employ physical adversarial attacks? For a hypothetical use-case to not be considered contrived, that use-case should satisfy the condition that the threat model does not support any other attack methods that are easier to implement and/or are more effective \cite{gilmer2018motivating}. For example, printing out a patch and adding it to a stop sign might induce a failure in a self-driving car, but so would covering up the stop sign with a plastic sheet, or even simply knocking the sign down – although, to be fair, these attacks would change the semantic content of the scene and would likely fool some fraction of humans. Whether such drastic changes are allowable will depend on the threat model. 

These questions motivated the current work. We carried out a series of experiments involving adversarial patch attacks against object detection systems. Our primary focus was to investigate the robustness of the attacks under more challenging test conditions than are generally considered in the literature. Specifically, we considered attacking object detection models trained to recognize vehicles from aerial or satellite images. Compared to past experiments, these scenarios featured a large variation of the scale factor or relative size of the target vehicle within the image, as well as a greater range of camera angles and lighting conditions. Our experiments uniformly pointed to the conclusion that patch attacks against overhead object detection models are rather brittle, in that it is very difficult to successfully transfer the attacks from digital to physical settings. Furthermore, by experimenting under real-world conditions, we highlight practical challenges needed to implement the attacks, including issues with data collection and the insertion of the physical patch into new scenes. We caution the reader that our investigations were not comprehensive; there is a vast range of parameters and experimental settings that could be varied, and we did not make any attempt to systematically explore this space. Also, some of our evaluations were qualitative, rather than quantitative. These limitations (which are discussed in more detail below) notwithstanding, the main conclusion of this empirical study is the observation that adversarial examples do not pose a serious threat to overhead object detection models, at least not with current models and attack methodologies.

\section{Related Work}
There is by now a large body of work on adversarial examples, which were first studied in the context of perturbations to neural networks in \cite{szegedy2013intriguing}. Initial research considered digital attacks; physical attacks were first investigated in \cite{kurakin2016adversarial}. The first adversarial examples involved slightly perturbing all the pixels of an image, which is not well-suited for practical implementations of this attack in the physical world. \cite{brown2017adversarial} introduced adversarial patches for object classification systems. Compared to previous attacks, the modified pixels in the adversarial patch attack are constrained to lie in some localized region (the patch), and moreover the modified pixels are not constrained to be close to their original values. This attack was extended to object detection models in \cite{liu2018dpatch} and further improved in \cite{lee2019physical}. Other attacks against object detection models include \cite{braunegg2020apricot, lu2017no, chen2018shapeshifter, thys2019fooling, wu2020making, xu2020adversarial, wang2021towards}. Lastly, \cite{gilmer2018motivating} critically considered the threat models used in papers studying the security threat posed by adversarial examples and concluded that many of the threat models are unrealistic or contrived. 

\section{Experimental Set-Up}
In this section we describe the experimental set-up, including the test-site, filming details, model architecture, and model training set.

\subsection{Experimental Test Site}
\begin{figure}
    \centering
    \includegraphics[width=0.5\textwidth]{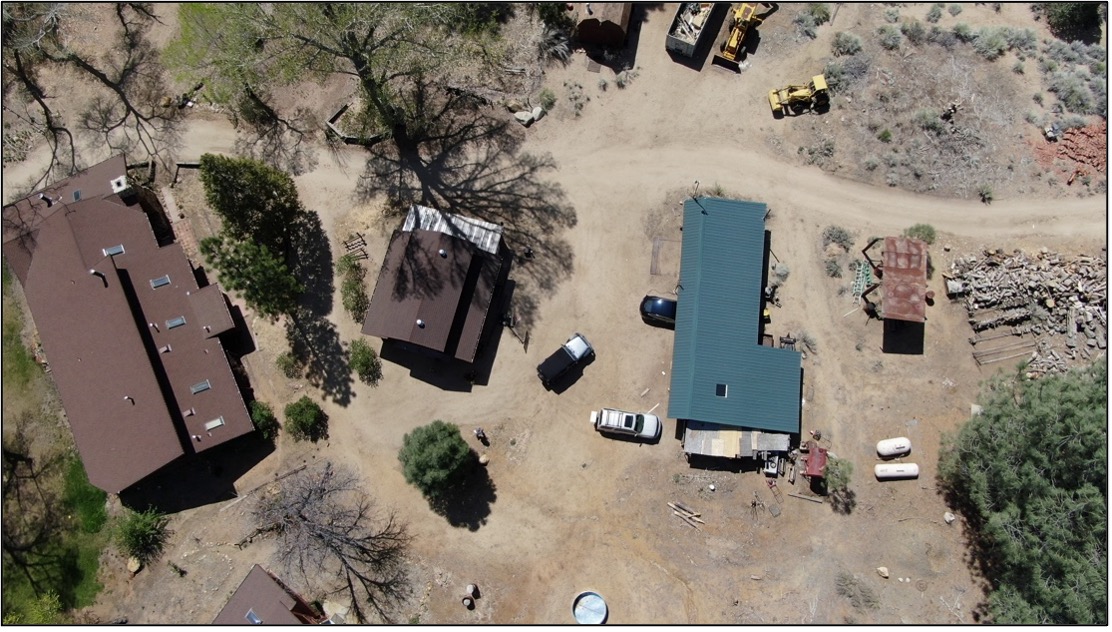}
    \caption{Aerial shot of the Mojave property. Three cabins are shown (the buildings with brown roofs), as well as the garage (building with the turquoise-colored roof). This shot also depicts multiple vehicles, including the Jeep Wrangler (located in the center of the picture). Also shown are various items such as gas tanks, construction vehicles, loose rubble, and cleared wood/brush.}
    \label{fig:mojave}
\end{figure}

For the test site we acquired access to a property located in Kern County, California. The property contained multiple buildings and structures, including cabins, a garage, and a shed, and the terrain was populated with typical desert foliage, including Joshua and pine trees, as well as smaller brush, see Fig.~\ref{fig:mojave}.

\begin{figure}
    \centering
    \includegraphics[width=0.5\textwidth]{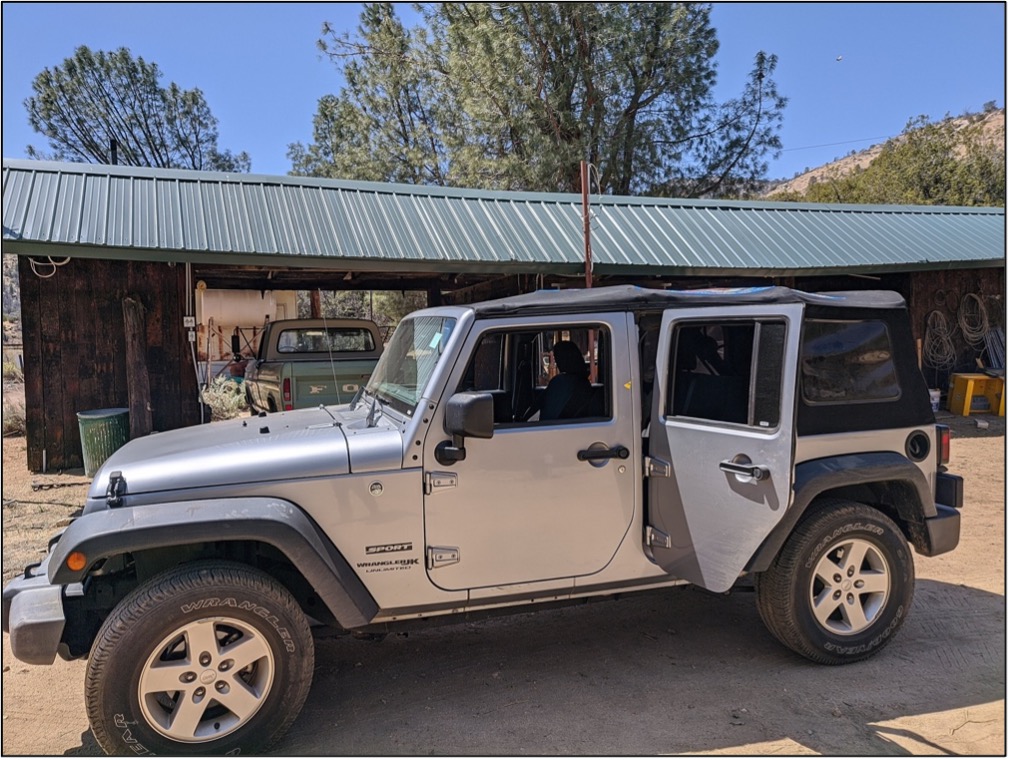}
    \caption{The Jeep Wrangler used for the experiment.}
    \label{fig:jeep}
\end{figure}

We also used a vehicle (a Jeep Wrangler, Limited Edition) to be used as the subject for the filming, as shown in Fig.~\ref{fig:jeep}. This test site represents a challenging environment for both the object detection machine learning model and the adversarial attack, and is therefore a good testing bed to study operational implications. For example, the Sun was quite bright on the testing day, and this caused glares on the Jeep and other reflective surfaces, such as the other vehicles located on the property, or the glass and metallic materials used in the construction of the buildings. Additionally, as the Jeep drove throughout the property it kicked up clouds of loose dirt and it also became partially occluded by trees and other foliage. These environmental effects obstructed the ability of the ML model to successfully identify the Jeep and they also distorted the raw pixel values of the adversarial patch.

We hired a drone photographer to take high-resolution aerial footage of the Jeep driving throughout the property, both with and without adversarial patches present in the scene. Video footage was taken (as opposed to still photographs), and 22 different video sequences were acquired, representing about 42 GB of data. In each sequence the drone trailed the Jeep as it drove around the property. Most of the footage was taken at an altitude of 400 ft, although we also filmed the drone descending from or ascending to this altitude, and some shots also feature the drone “swooping in” by descending while moving horizontally towards the vehicle. Lastly, all but one of the 22 video sequences was filmed in 1080p (corresponding to a resolution of {$1920\times1080$}); the other was filmed at 4K (corresponding to a resolution of {$3840\times2160$}).

\section{Target Model Selection}
To assess the efficacy of an adversarial attack, a target ML model is required. While adversarial attacks exist for many ML models acting upon many data modalities, we decided to focus on object detection models. For the assessment to be fair, it is important that the target models used be reasonably well-trained and appropriate for the circumstances. This is admittedly a rather vague and subjective criterion. In our case, our goal was to attack models that were trained using a reasonable amount of compute and engineering resources. For example, we elected to use the well-known and widely implemented model known as Faster RCNN (FRCNN) \cite{ren2015faster}. (An interesting class of models that we did not explore are edge models, which are designed to operate with low-power requirements, typically through the use of comparatively smaller neural networks and quantized parameter values.) We are not aware of any publicly available model ideally suited for this environment, nor did we find an obvious labeled dataset which could be used to train such a model. We therefore considered four versions of the model, one pre-trained on the Microsoft COCO dataset \cite{lin2014microsoft}, and three models which were further fine-tuned on relevant publicly-available datasets of aerial or satellite imagery. As described below, each of these datasets contains some, but not perfect, similarity to the testing conditions: the xView datasets consists of satellite images; the COWC (Cars Overhead With Context) dataset consists of images taken from aircraft; and the OATML (Operationally-relevant Artificial Training for Machine Learning) datasets consists of images taken from a low-flying drone.  

None of the four models are state-of-the-art, nor is any one optimized for the testing conditions. This choice was motivated in part by practical considerations as well as our view that many realistic use-cases will feature either pre-trained or lightly fine-tuned ML models being used in conditions that correspond to a notable deviation from the training set distribution. Of course, as a result of this experimental design choice we will not be able to make any claims about heavily optimized models tailored for very specific settings. We discuss these three fine-tuned datasets next; the technical details of the fine-tuning procedure are discussed in the Appendix.

\subsection{Fine-Tuning Dataset: Xview}
The first fine-tuning dataset we considered was the dataset associated with the DIUx xView 2018 Detection Challenge \cite{lam2018xview}. This dataset was assembled by the Defense Innovation Unit Experimental (DIUx) and the National Geospatial-Intelligence Agency (NGA) to spur the development of new solutions for both national security and disaster response. The xView dataset contains satellite images of complex scenes from around the world with 60 object classes including “cargo plane”, “passenger vehicle”, “truck”, “maritime vessel”, “building”, and “storage tank”, to name a few. The appeal of the xView dataset for our current purposes is the fact that the images are taken from a top-down perspective and contain complex urban and industrial scenes with a range of vehicle and building types identified. 
One minor technical detail is that the images are rather large, and as a preprocessing step we split them into {$512\times512$} resolution “chips”.\footnote{“Chipping”, the process of splitting a large image into smaller sub-images, has little effect on the predictions of the FRCNN model which considers many proposed regions when processing an image. The main effect of the chipping is to restrict the proposed regions to those that entirely lie within a single chip.}  A model trained on the xView dataset would excel at identifying objects from very high altitudes but would probably not do well at closer ranges. The xView images have a reported resolution of {30 cm $\times$ 30 cm/pixel}, according to the DIUx xView challenge website.\footnote{\url{http://xviewdataset.org/\#dataset}} \footnote{It is important to note that the term ‘resolution’ has two distinct colloquial interpretations in the current context. It may refer to the pixel resolution, i.e., the number of pixels in the image (for example, a 1080p resolution image consists of $1920\times1080$ pixels), or it may refer to the physical scale of the image (for example each pixel in the xView images corresponds to an area of 30 cm $\times$ 30 cm). We will consider both concepts in the following.} 

\subsection{Fine-Tuning Dataset: COWC}
The second fine-tuning dataset we considered was the Cars Overhead With Context (COWC) dataset \cite{mundhenk2016large} from Lawrence Livermore National Laboratory. This dataset is dedicated to the detection of vehicles, and there are just four object classes: “sedan”, “pickup”, “other”, “unknown”. This dataset is therefore relevant for applications where the only objects of interest are vehicles. 
The COWC images have a reported resolution of {15 cm x 15 cm/pixel}. 

\subsection{Fine-Tuning Dataset: OATML}
The third and final fine-tuning dataset we considered was developed during a previous RAND project designed to test whether neural networks trained on frames extracted from video games could perform well on real-world images \cite{hartnett2020operationally}. Although most of the images collected during this project were from the video game ARMA 3, for evaluating purposes a second dataset was created of a rented High Mobility Multi-purpose Wheeled Vehicle (HMMWV) in the California desert, and it is this dataset that we used to fine-tune the pre-trained FRCNN model. In this prior work, the drone took both a series of high-definition photographs and video footage in 1080p resolution. The {$4000\times3000$} resolution photographs were used for the training set, and {$1920\times1080$} resolution frames extracted from the video were used for the test set. Although there is only a single object category in this dataset, “HMMWV”, it has many similarities to the images collected for the present project because the same equipment, operator, and filming style were used. The acronym OATML is used to refer to this dataset (for Operationally-relevant Artificial Training for Machine Learning). 
A model trained on the OATML images would likely excel at identifying images from relatively close ranges or low altitudes.

\subsection{Summary}
In total, we considered attacks against four distinct models – one for each of the three fine-tuning datasets, and we also decided to attack the pre-trained model itself, i.e., the model with no fine-tuning.\footnote{The test split of the COWC dataset was used to generate adversarial attacks against the pre-trained model. This required that we map the COWC labels to COCO labels, which we did as follows: COWC:sedan $\rightarrow$ COCO:car, COWC:other $\rightarrow$ COCO:car , COWC:unknown $\rightarrow$ COCO:car, , COWC:pickup $\rightarrow$ COCO:truck.} The full set of models considered is listed in Table 2.1. We note that none of these models are perfectly aligned with our experimental needs, in that the training images will differ in a number of ways from the images we use to test the models and attacks on. Of course, for the sake of improving model performance and isolating the effect of the adversarial attacks, it would be desirable to use images from the same dataset for both training and testing. However, as mentioned above, we view the current situation as more representative of many practical applications of ML systems. Therefore, the fact that each of the fine-tuning datasets is only roughly like the data collected in the experiment is actually desirable from our perspective.

\begin{table*}[htbp]
\caption{\label{table:modeltable} Target models used in the experiments.}
\centering
\begin{tabular}{c|cccc}
    & Model 1 & Model 2 & Model 3 & Model 4 \\
    \midrule
    pre-training dataset & COCO & COCO & COCO & COCO \\
    fine-tuning dataset & N/A & xView (train) & COWC (train) & OATML (train) \\
    evaluation dataset & COWC (test) & xView (test) & COWC (test) & OATML (test) \\    
\end{tabular}
\end{table*}

\section{Adversarial Attack}
We considered attacking the four FRCNN models described above using the Lee-Kolter \cite{lee2019physical} adversarial patch attack, which expands upon the earlier DPatch attack by Liu \textit{et al} \cite{liu2018dpatch}. 
We chose to consider the Lee-Kolter attack because it was previously demonstrated to perform well at attacking object detection models, both digitally and in the physical world. Both attack methods can be seen as extensions of the original adversarial patch attack \cite{brown2017adversarial}. The original attack applied to image classification models, whereas both the Liu \textit{et al.} and Lee-Kolter attacks apply to object detection models such as FRCNN.

Briefly, the Lee-Kolter attack works as follows: first, a patch size and location are selected, which defines the patch region within the image. Only pixels within the patch will be modified by the attack; all others will be left unchanged. The pixel values of the patch are then varied to maximize the loss of the object detection model – it is in this sense that the patch is ‘adversarial’ as this optimization works against the objective used to train the model. The object detection loss function depends on both the input image x as well as the targets y (which consist of a set of class labels and bounding boxes).\footnote{Note that the word target here refers to the combination of bounding box coordinates and object classes, symbolically represented by a collection of tuples y. We also occasionally use the term target to refer to the main object of interest within an image.} The targets used in the attack are those corresponding to the model predictions on a clean, or un-attacked image – thus the attack is working to counter-act the predictions of the un-attacked model rather than to counter-act the ground truth targets. This is an important distinction as it allows the attack to be used even when the ground truth labels are not known.\footnote{If the model is very well trained, the ground truth labels and model predictions should be similar, but in practice we can expect that the model predictions will deviate significantly from the ground truth targets.}  An important feature of the Lee-Kolter attack is that during the optimization loop of the patch pixel values, the input image x is sampled from a dataset, and a random transformation (comprised of a random rotation, scaling, translation, and brightness adjustment) is applied. This “Expectation Over Transformation” method \cite{athalye2018synthesizing} helps ensure the generated patch is robust, and not tied to a specific image or scene configuration. Finally, the pixel values of the image are clipped to lie within the acceptable range for images, which is a subtle but important step if the attack is meant to transfer to the physical world.
These steps are then repeated many times in an outer optimization loop.

\subsection{Threat Model}
We considered both white and black-box adversarial patch attacks against the FRCNN object detection model created using the Lee-Kolter algorithm \cite{lee2019physical}. Our primary focus was on physical attacks, wherein adversarial patches are inserted into real-world scenes that are then captured on camera. To construct these attacks, the patches must first be generated against existing images, which corresponds to a digital attack. The digital attacks were white-box, and the physical attacks were both white-box and black-box, as we applied all four models to every image containing a physical patch (with each patch having been generated from one of the four models). The FRCNN models were pre-trained on COCO and then fine-tuned on various datasets of overhead images. We assumed that the attackers also had access to the fine-tuning datasets, which were used to generate the digital attacks. In contrast, the physical attacks were carried out under conditions that differed noticeably from the training images. Our threat model thus represents less-than-ideal conditions for both the attacker and defender in that the defending model was not highly optimized for the experiment location and in the fact that both players must contend with the distributional shift in going from the fine-tuning data to the test environment. The goal of the attacker is to fool the object detection algorithm by either causing false positives (decoys) or false negatives (camouflage). Of course, low-tech and low-cost solutions for each of these failure modes already exist, and a successful demonstration of the adversarial attack would be most interesting if it outperformed these existing approaches. For example, a successful adversarial patch attack should be more effective than an attack based on print-outs of other objects or textures that might fool the model. We did not consider such attacks here (however, this concern is not terribly relevant for the current work since our physical attacks performed so poorly, as we will demonstrate shortly).

\subsection{Implementation Details}
We used the Adversarial Robustness Toolbox software library \cite{nicolae2018adversarial} to implement both the target FRCNN models as well as the adversarial patch attacks.\footnote{The Lee-Kolter attack is referred to as the “Robust DPatch” attack, as compared to the DPatch attack of \cite{liu2018dpatch}.}  For each of the four models considered we produced two adversarial patches, a smaller one ({$160\times160$ pixels}) and a larger one ({$320\times320$ pixels}). Aside from the size, the training details were equivalent, with one exception described below. Each patch was produced using 150,000 iterations of projected gradient descent applied to the FRCNN loss function with a learning rate of 10.0 (the {$320\times320$} OATML patch is an exception, where training was halted after 88,000 iterations). Each iteration used a different base image for the attack, so that throughout the course of the attack generation all the images in the training set were used. The Lee-Kolter attack allows for image transformations to be sampled throughout the training process (that is, the Expectation over Transformation method). We did not make use of this functionality, except during the training of one patch, the {$160\times160$} xView patch. In this case, we added a random resizing transformation with a scale factor uniformly drawn from the interval [0.5, 2] in the inner optimization loop to improve the robustness of the attack to different scales. Finally, all the deep learning aspects of the computer vision experiments were coded using PyTorch and run on a single desktop computer equipped with an Nvidia GTX 1080 Ti GPU.

To test the adversarial patches in the physical world, we printed them on a variety of paper products. Because the attack efficacy could depend on the patch size and the quality of the paper, we considered a range of printing methods. For our main experiments, we printed two patches on {4 ft $\times$ 4 ft} square pieces of matte paper. We also considered patches printed on standard A5 office paper, {2 ft $\times$ 3 ft} paper, and {2 ft $\times$ 3 ft} fabric paper. To convert the raw pixels of the patch into a printable image, we saved the patches in .png format with a dpi (dots per inch) value of 900. Lastly, we note that the effect of the printing details on the patch performance is an important issue worthy of a more systematic investigation. 

\section{Results}
In assessing the performance of the attack, it is reasonable to separately consider attacks acting digitally and attacks acting in the physical world. 
Naturally, one can expect that the digital attacks will be much more effective, and we will find that this expectation is born out in practice. In fact, we found it extremely difficult to demonstrate a successful physical attack even when the digital attack worked well.

\subsection{Digital Attacks}
A natural measure of the efficacy of the digital adversarial patch attack is how the mAP and mAR (mean average precision and mean average recall) decrease when the patch is added to images in the testing set. This is shown in Table 4.1 for all four models and both patch resolutions. 
Both the mAP and mAR scores were computed using the COCO evaluation standard, with IoU=0.50:0.95, area=all, maxDets=100.\footnote{Here, IoU (intersection over the union) is a measure of the overlap between the predicted and ground truth bounding box, ranging from 0 (no overlap) to 1 (total overlap). The mAP/mAR scores were computed using different IoU detection thresholds, ranging from 0.5 to 0.95. For more details on the COCO evaluation standard, see \url{https://cocodataset.org/\#detection-eval}.}  

Of course, some of the mAP/mAR decrease is simply because the patch obscures a significant fraction of the image. To measure the extent of this effect, we also include the mAP/mAR values for a randomly initialized patch added to the image. Unsurprisingly, the general trend is that adding a randomly initialized patch to the image causes performance to drop, and then a further drop is facilitated by the optimization associated with the adversarial generation of the patch. Also unsurprisingly, larger patches are more effective than smaller patches.

Model 1 (which was not fine-tuned) performs abysmally on each test set, including the unattacked clean images. This is unsurprising because this model was trained on COCO images, and the test sets consisted of overhead COWC images. The results for the other models behave as expected; the highest scores are achieved on the clean images, the second highest scores are achieved on the images with randomly initialized patches inserted, the next highest score corresponds to the 160x160 trained patches and finally the worst score corresponds to the 320x320 trained patch. An important observation is that, judged simply by the mAP/mAR score, the OATML model performs best, followed by the COWC model and then by xView model (which achieves quite low scores).\footnote{The 0.05 mAP score of our model should be compared against the baseline model released by the competition organizers which achieved an mAP of about 0.22, and against the winning model which achieved a score of 0.28 \cite{sergievskiy2019reduced}.} This ranking most likely reflects the relative difficulties of each dataset, for example the OATML data has a single class, the COWC data has four, and the xView data has 60. Therefore, care should be taken in directly comparing the scores of one model against the scores of another. 

\begin{table*}[htbp]
\caption{\label{table:attacktable} Test set mAP/mAR values .}
\centering
\begin{tabular}{c||cccc}
     & Model 1 & Model 2 & Model 3 & Model 4 \\
    Attack $\backslash$ Fine-Tuning & None & xView & COWC & OATML \\
    \midrule \midrule
    None & 0.00/0.00 & 0.05/0.12 & 0.45/0.62 & 0.71/0.77 \\ \hline
    160x160 (initialized) & 0.00/0.01 & 0.04/0.11 & 0.40/0.56 & 0.71/0.76 \\
    160x160 (trained) & 0.00/0.00 & 0.00/0.00 & 0.19/0.49 & 0.14/0.74 \\ \hline
    320x320 (initialized) & 0.00/0.00 & 0.03/0.07 & 0.27/0.38 & 0.71/0.76 \\
    320x320 (trained) & 0.00/0.00 & 0.02/0.03 & 0.00/0.01 & 0.02/0.73 \\    
\end{tabular}
\end{table*}

\begin{figure*}
    \centering
    \includegraphics[width=1.0\textwidth]{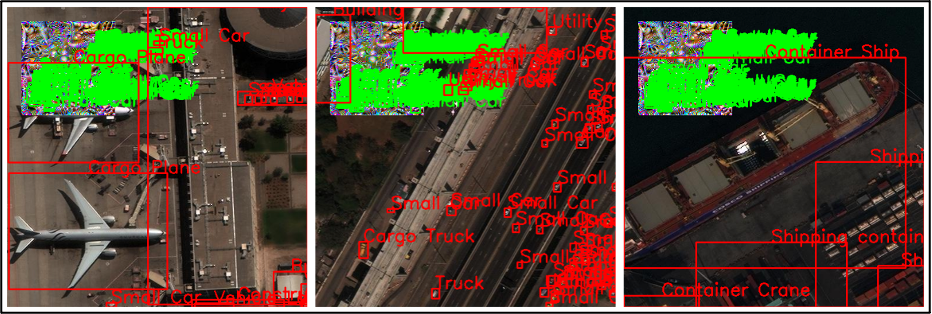}
    \caption{The 160x160 resolution adversarial patch digitally applied to test set xView images.
    }
    \label{fig:xview_patch}
\end{figure*}

\begin{figure*}
    \centering
    \includegraphics[width=1.0\textwidth]{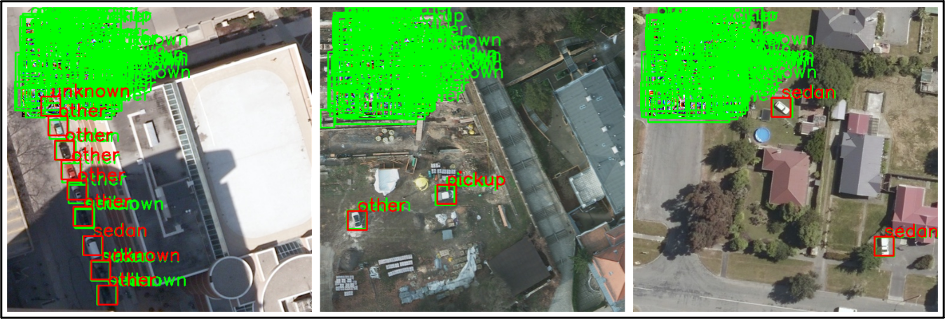}
    \caption{The 160x160 resolution adversarial patch digitally applied to test set COWC images. 
    }
    \label{fig:cowc_patch}
\end{figure*}

\begin{figure*}
    \centering
    \includegraphics[width=1.0\textwidth]{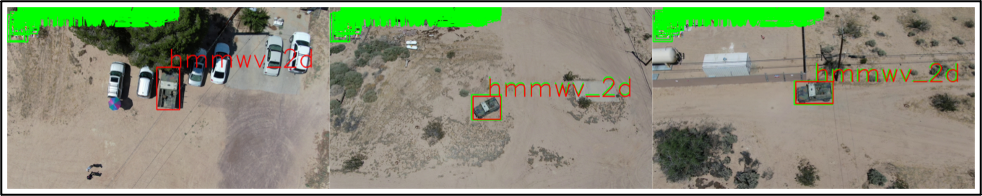}
    \caption{The 160x160 resolution adversarial patch digitally applied to test set OATML images 
    }
    \label{fig:oatml_patch}
\end{figure*}

It is useful to inspect a few examples to gain a sense for the effect the patches have on the models’ predictions. These are shown in Fig.'s~\ref{fig:xview_patch}, \ref{fig:cowc_patch}, \ref{fig:oatml_patch} for the case of the xView-trained model (i.e. Model 2) applied to xView, COWC, and OATML images, respectively. The model predictions are shown in green in order to distinguish them from the ground truth bounding boxes, which are shown in red. It engenders many false detections within the patch area (false positives), and it also suppresses detections of real objects that exist outside of the patch area (false negatives). The patch therefore acts as a kind of decoy that draws attention from the real objects of interest in the scene. The patch is apparently less effective when applied to the COWC images, where it continues to cause many false positive detections, but it fails to suppress valid detections of objects outside the patch. When applied to the OATML images, the patch once again causes many false positives, and in some, but not all, of the images it also suppresses detection of the HMMWV. Interestingly, the false positive bounding boxes can be seen to extend outside the patch area in the xView and OATML images, with the effect rather dramatic for the OATML images. Conversely, the false positive bounding boxes for the COWC images seem rather localized to the patch area.

Further insight into the attack performance can be gained by inspecting the optimization problem used to develop the attack. The pixel values of the adversarial patch are generated by maximizing the FRCNN loss, which is itself a combination of four sub-losses \cite{ren2015faster},

\begin{equation}
	\ell = \ell_{\text{classifier}} + \ell_{\text{box reg.}} + \ell_{\text{objectness}} + \ell_{\text{RPN box reg.}} \,.
\end{equation}

Each sub-loss measures one aspect of the overall detection problem. Two of the losses relate to the predicted class: $\ell_{\text{classifier}}$ is the standard classification loss measuring how well the ground truth label compares with the predicted label probabilities and $\ell_{\text{objectness}}$ is a binary classification loss which simply measures whether the prediction agrees with the presence or absence of an actual object, irrespective of the specific class. The other two losses, $\ell_{\text{box reg.}}$, $\ell_{\text{RPN box reg.}}$, evaluate the quality of the predicted bounding box. 

\begin{figure}
    \centering
    \includegraphics[width=0.5\textwidth]{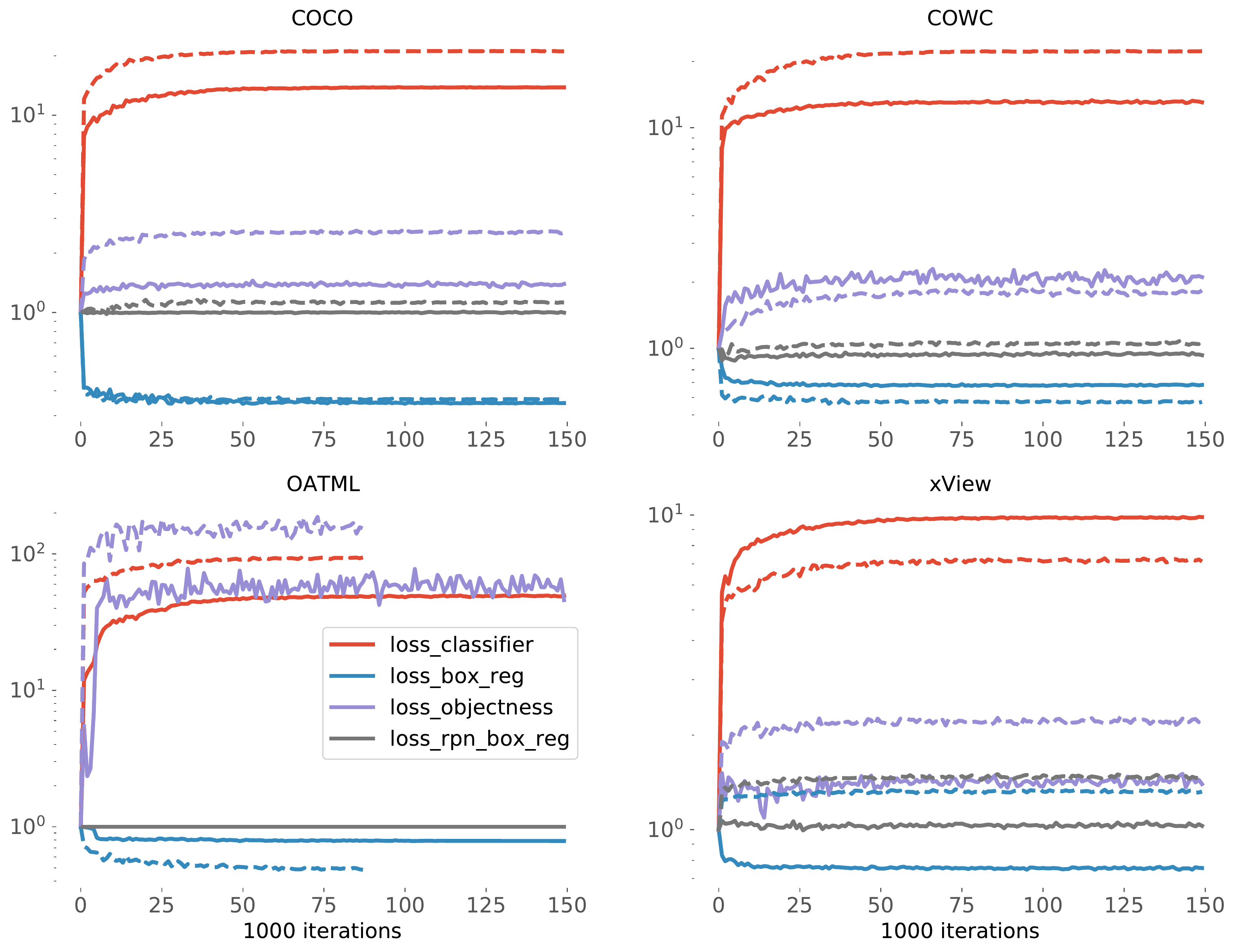}
    \caption{The relative changes in the four FRCNN losses through the training of the adversarial patch, i.e. $\ell_t/\ell_0$ for $t = 0, 1, ..., T-1$ the training iteration and $\ell_t$ one of the four losses. Each subplot shows two sets of results for each of the four losses, with the solid curves representing the 160x160 resolution patches and the dashed curves representing the 320x320 resolution patches.}
    \label{fig:losses}
\end{figure}

In Fig.~\ref{fig:losses} we plot the relative change in these losses over the course of the patch training for both patch sizes, {$160\times160$} and {$320\times320$}. (The relative change, as opposed to the absolute change, is plotted because the typical values of these sub-losses can differ by a few orders of magnitude.) Each subplot shows two sets of results for each of the four losses, with the solid curves representing the {$160\times160$} resolution patches and the dashed curves representing the {$320\times320$} resolution patches. These plots reveal that the loss most significantly impacted by the presence of the patch is the classifier loss. The objectness loss, which measures whether the predictions correspond to real objects, also increased substantially throughout training. The OATML-trained model is unique in that the objectness loss is comparable to the classifier loss; for all other models the objectness loss is roughly half an order of magnitude smaller. The OATML result is likely because the data has just a single class label. Interestingly, the bounding box losses behave quite differently from the classification/objectness losses. The RPN bounding box regularization loss is more or less unaffected by the presence of the patch, and the bounding box regularization loss actually decreases as the patch is trained.

\subsection{Physical Attacks}
We considered two ways of inserting the printed adversarial patches into the scene: 1) we attached the patch to the Jeep and had the drone track the Jeep as it drove throughout the property, and 2) we placed various patches on the ground along the route of the Jeep. If the patch works in the physical world as it does digitally, then we can expect that the patch will engender many false positive detections localized inside the patch, and it will also suppress detections of neighboring objects (i.e., it will create false negatives). If the goal of the attack is to suppress detection of the Jeep, then it would not make practical sense to attach the patch to the Jeep itself. This is because even if the patch suppressed actual detection of the Jeep, it would create multiple false positives centered on the Jeep, in effect replacing one type of detection with another. However, for the purposes of this experiment we found it convenient to affix the patch to the Jeep so that a variety of background scenes could be explored as the Jeep traversed the property.

\begin{figure}
    \centering
    \includegraphics[width=0.5\textwidth]{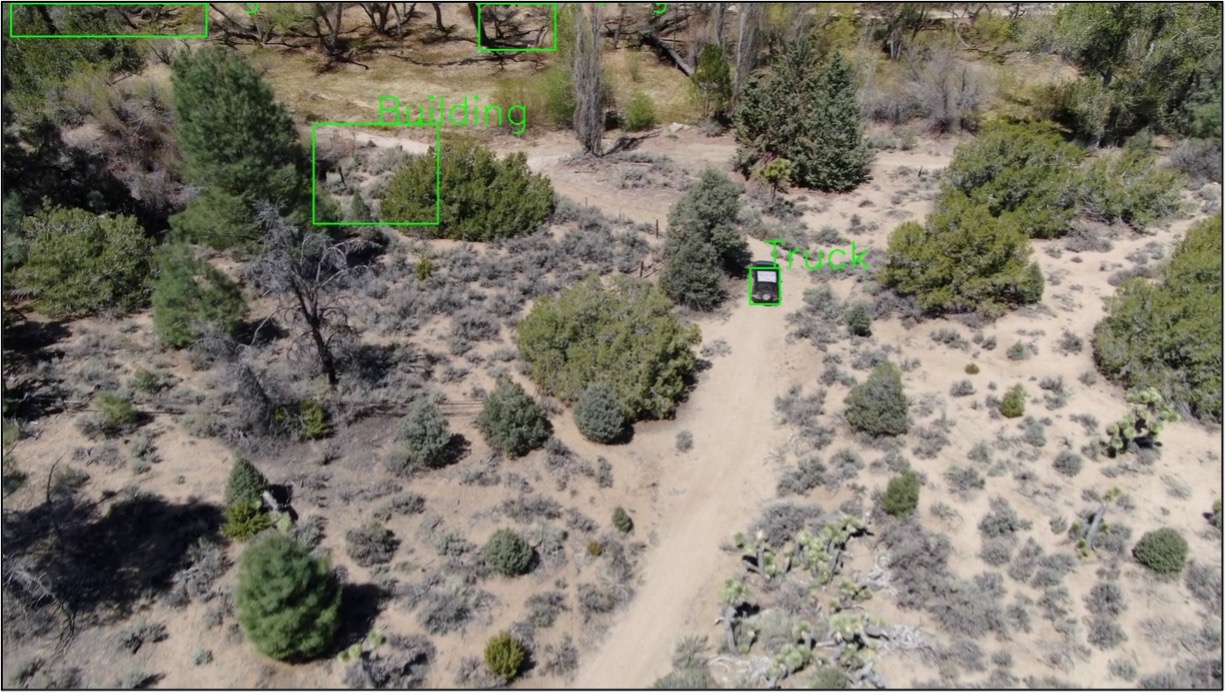}
    \caption{
    Example frame showing the predictions of the xView model, viewed at close range. The 4 ft x 4 ft, 320x320-resolution adversarial patch is affixed to the roof of the Jeep has no apparent effect on the predictions: there are no false positives localized on the patch, and the (spurious) detections nearby were not prevented. The misclassifications of the model reflect the fact that this model was trained on satellite images with a much lower scale factor.}
    \label{fig:physical_example3}
\end{figure}

We knew that our testing conditions would be a challenge for the physical adversarial patches, and we expected that the attacks might fail once the drone ascended past some threshold altitude and the captured images of the patches became smaller. Unfortunately, getting the patches to work in the physical world at all turned out to be harder than we anticipated, and we were unable to identify a single frame in which the patches worked as intended. In other words, the adversarial patch completely failed our test. Some example frames that include the Jeep with these adversarial patches attached are shown below in Fig.'s \ref{fig:physical_example1}, \ref{fig:physical_example2}, \ref{fig:physical_example3}. In all cases the 4 ft $\times$ 4 ft adversarial patch that was affixed to the roof of the Jeep had no apparent effect on the predictions: there are no false positives localized on the patch, and the detection of the Jeep (as a haul truck) was not prevented. (In Fig.~\ref{fig:physical_example3} the Jeep is incorrectly classified as a building: this is likely due to the close-up scale of the image.) The images also show many other detections of the model. Many of the predicted classes are reasonable given that the model was trained on the xView dataset and not images from similar desert scenes. We observed this same failure for each of the patch print-outs we tested, which encompassed a range of models, patch resolutions, print-out size, and paper material. 

\begin{figure}
    \centering
    \includegraphics[width=0.5\textwidth]{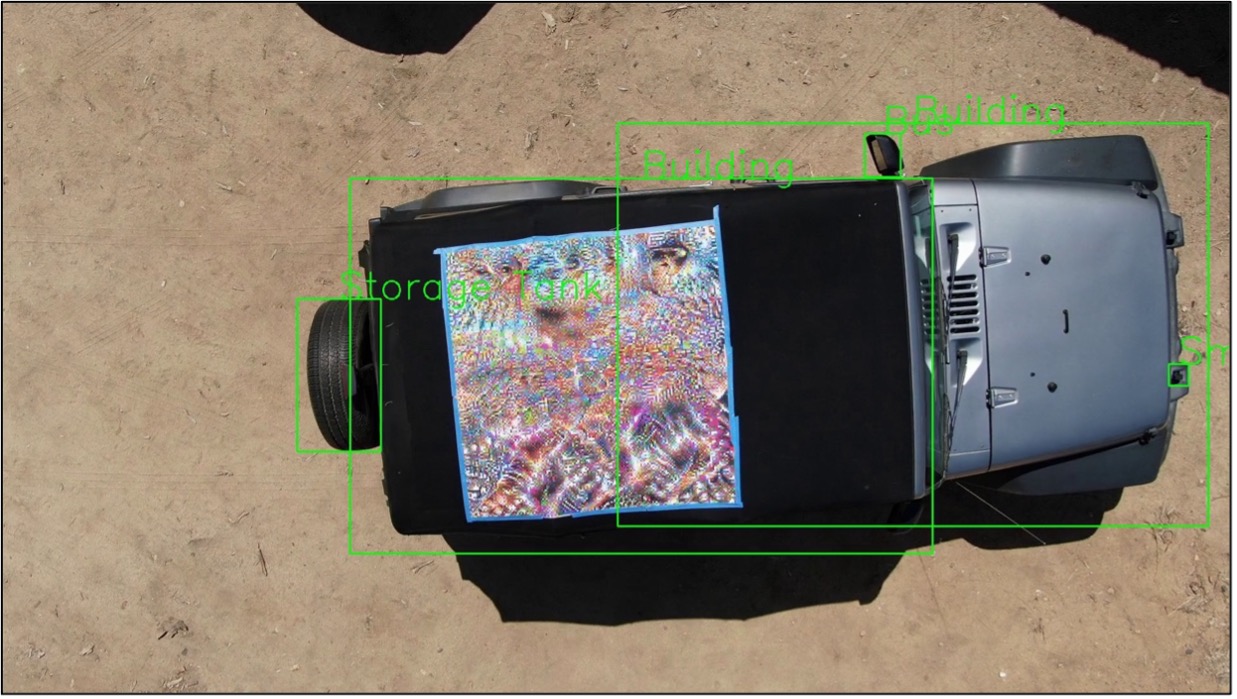}
    \caption{
    Example frame showing the predictions of the xView model. The 4 ft x 4 ft adversarial patch is affixed to the roof of the Jeep has no apparent effect on the predictions: there are no false positives localized on the patch, and the detection of the Jeep (as a haul truck) was not prevented. The image also shows many other detections of the model. Many of the predicted classes are reasonable given that the model was trained on the xView dataset and not images from similar desert scenes.}
    \label{fig:physical_example1}
\end{figure}

\begin{figure}
    \centering
    \includegraphics[width=0.5\textwidth]{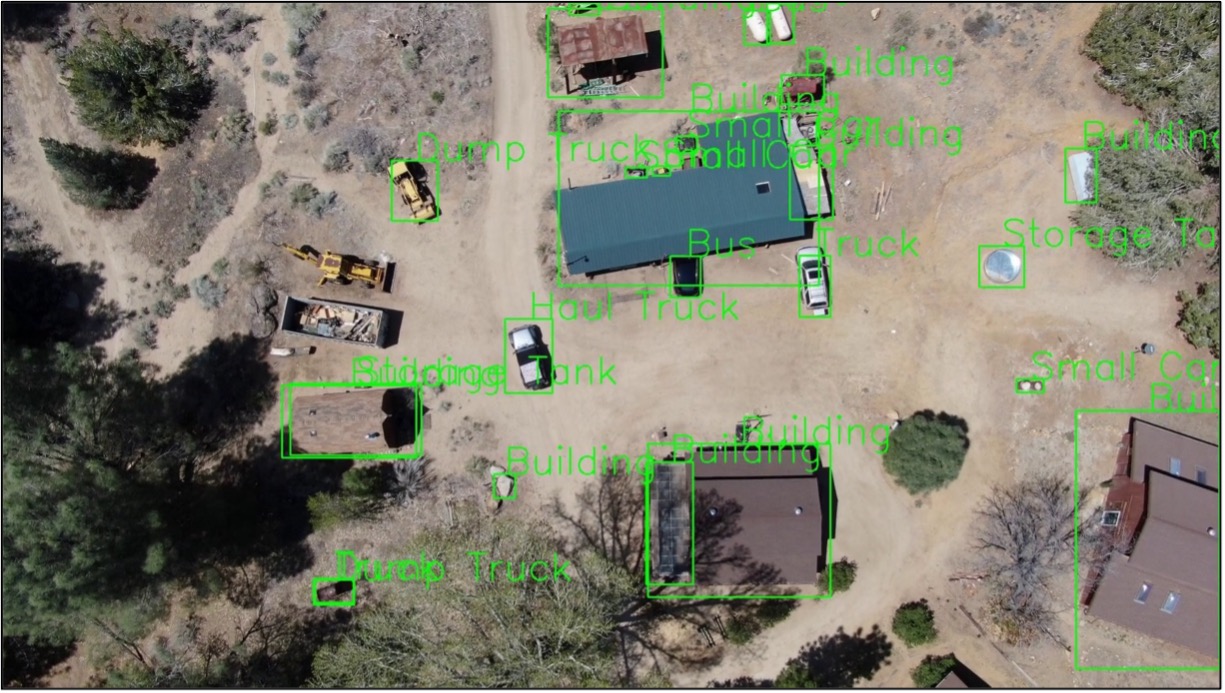}
    \caption{
    Example frame showing the predictions of the xView model. The 4 ft x 4 ft adversarial patch is affixed to the roof of the Jeep has no apparent effect on the predictions: there are no false positives localized on the patch, and the detection of the Jeep (as a truck) was not prevented. There are also spurious detections in the background.}
    \label{fig:physical_example2}
\end{figure}

\begin{figure}
    \centering
    \includegraphics[width=0.5\textwidth]{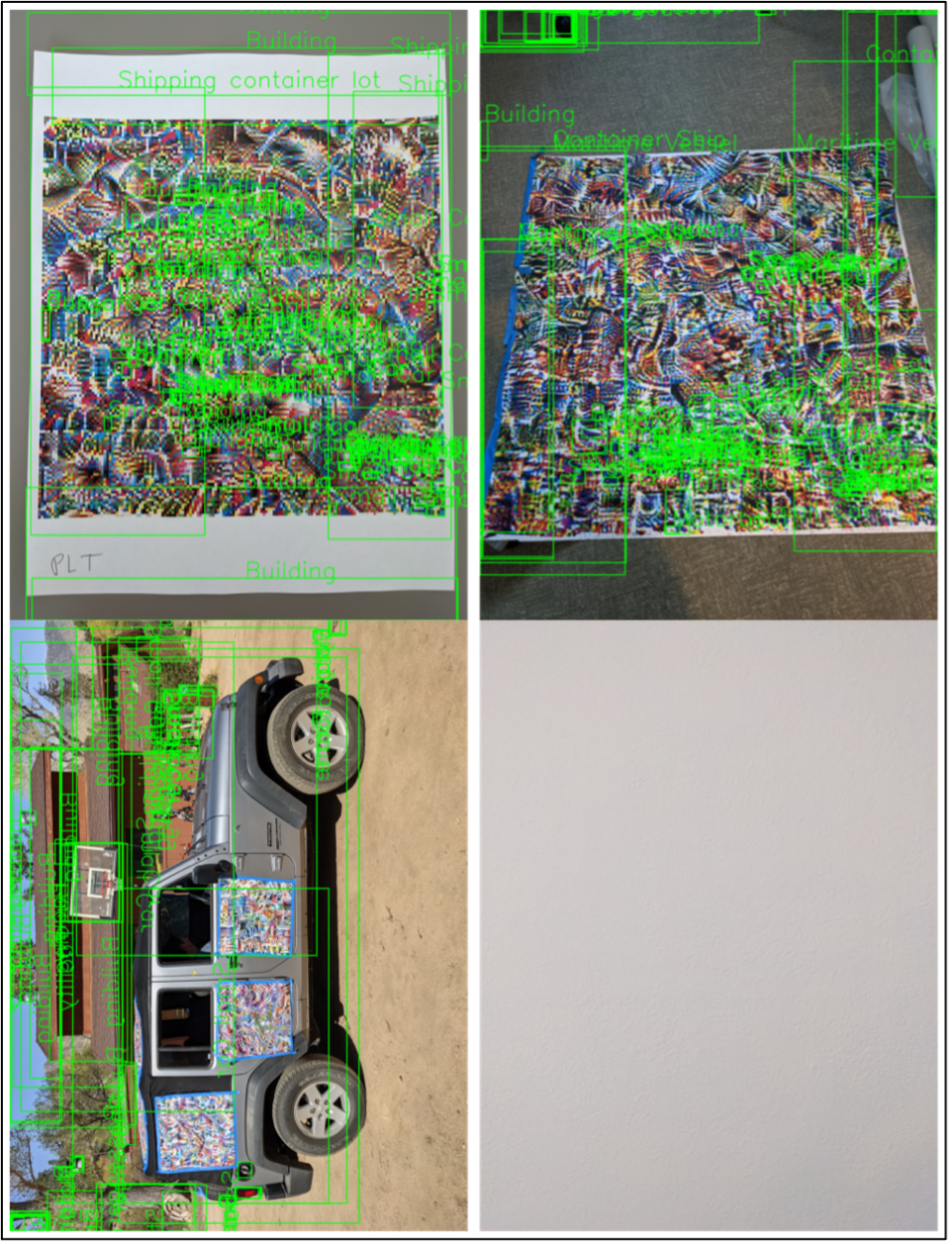}
    \caption{The FRCNN predictions for images of a patch printed on A5 paper (Top Left), 4 ft $\times$ 4 ft matte paper (Top Right), a cell-phone image of the Jeep (Bottom Left), and an image of a white wall (Bottom Right) as a sort of control image.}
    \label{fig:frcnn_preds}
\end{figure}

The rather dramatic failure of the patch attacks to transfer to the physical world could be the result of some sort of implementation error. We performed a check of this possibility by taking close-up pictures of the patches with cell-phone cameras and submitting the new images to the FRCNN models. The patches appear to be working, at least moderately well. Fig.~\ref{fig:frcnn_preds} depicts the FRCNN predictions for images of a patch printed on A5 paper (Top Left), and 4 ft $\times$ 4 ft matte paper (Top Right). Both images show the predictions of the xView-trained model (Model 2 in Table 1) on images of each patch. These patches are engendering a significant number of false positive detections. It is harder to assess whether the patches are also causing false negative errors by suppressing detections outside the patch area, although it does seem to be the case that there are some predictions of objects outside of the patch area.

Fig.~\ref{fig:frcnn_preds} also shows the model predictions for a cell-phone image of the Jeep, with multiple printed patches attached (Bottom Left). Not that the patches are not causing false positive detections within their area, unlike the above photos. This suggests that the scale and lighting differences could be the cause of the patch failure. Lastly, we also show an image of a white wall (Bottom Right) as a control image. The model did not predict any objects within this image, which confirms that the predictions in the other images are indeed caused by the objects and the background scene.

\begin{figure}
    \centering
    \includegraphics[width=0.5\textwidth]{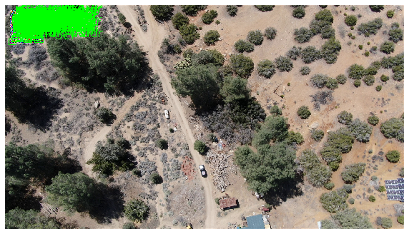}
    \caption{FRCNN predictions for a desert image with an adversarial patch digitally afixed.}
    \label{fig:desert_control}
\end{figure}

Overall, we found that the physical adversarial patches were not effective. Because the digital patches did perform well, the natural conclusion is that the failure was caused by the many challenges associated with going from a digital to a physical attack. However, one might object that such a conclusion is premature, because the digital and physical attacks were performed on two different sets of images; the digital attacks were performed against the xView, COWC, and OATML images, and the physical attacks were performed against the desert images. As a check, we verified that the digital attacks also work on the desert images, see Fig.~\ref{fig:desert_control} This lends further support to our conclusion that the main difficulty in executing the attacks was their transferrance to the physical world. 

\section{Patch Scale and Resolution}
A successful physical patch attack should work under a fairly broad range of real-world conditions. Ensuring that the patch attack remains successful for different camera orientations is especially difficult, as there are multiple scale transformations involved. First, there is a scale transformation associated with transferring the digital attack to the physical world. A second scale transformation is induced by varying the distance-to-camera of the printed-out patch. 

To ground this discussion in a concrete example, consider the $160\times160$ pixel patches trained on the xView dataset. 
The DIUx xView 2018 Detection Challenge website lists the pixel resolution of the xView images as {$0.3$ m}, translating to an area of {$(0.3 \times 160)^2 \text{ m}^2 = 2304 \text{ m}^2$} for the digital patch. On the other hand, the physical dimensions of the printed-out patch are set by the size of the paper - in the case of the the {4 ft $\times$ 4 ft} matte paper, this corresponds to a scale factor of roughly 40 for each dimension (height and width). 
Therefore, in transferring the patches from the digital to physical worlds we have implicitly made a dramatic scale transformation. 
The physical patches are then inserted into a scene and captured by a camera, resulting in a second physical pixel scale. Varying the distance-to-camera (as occurred when the Jeep moved relative to the drone in our experiments) then alters this scale, and a successful physical patch should be robust to such transformations. 


These considerations are especially important for overhead aerial or satellite images, and we suspect that the main reason our physical attacks performed poorly is that the patches were not sufficiently robust to these scale transformations. This hypothesis is difficult to test directly, but we did perform a simple set of experiments to confirm the extreme sensitivity of the adversarial patch to relative changes of scale. We considered both resizing the patch relative to the image (keeping the image fixed), as well as resizing the image relative to the patch (keeping the patch fixed). These experiments allowed us to investigate the effect of relative changes of scale on purely digital attacks - presumably any degradation in performance would only be exacerbated if the attacks were to be carried out in the physical world.

\begin{figure}
    \centering
    \includegraphics[width=0.5\textwidth]{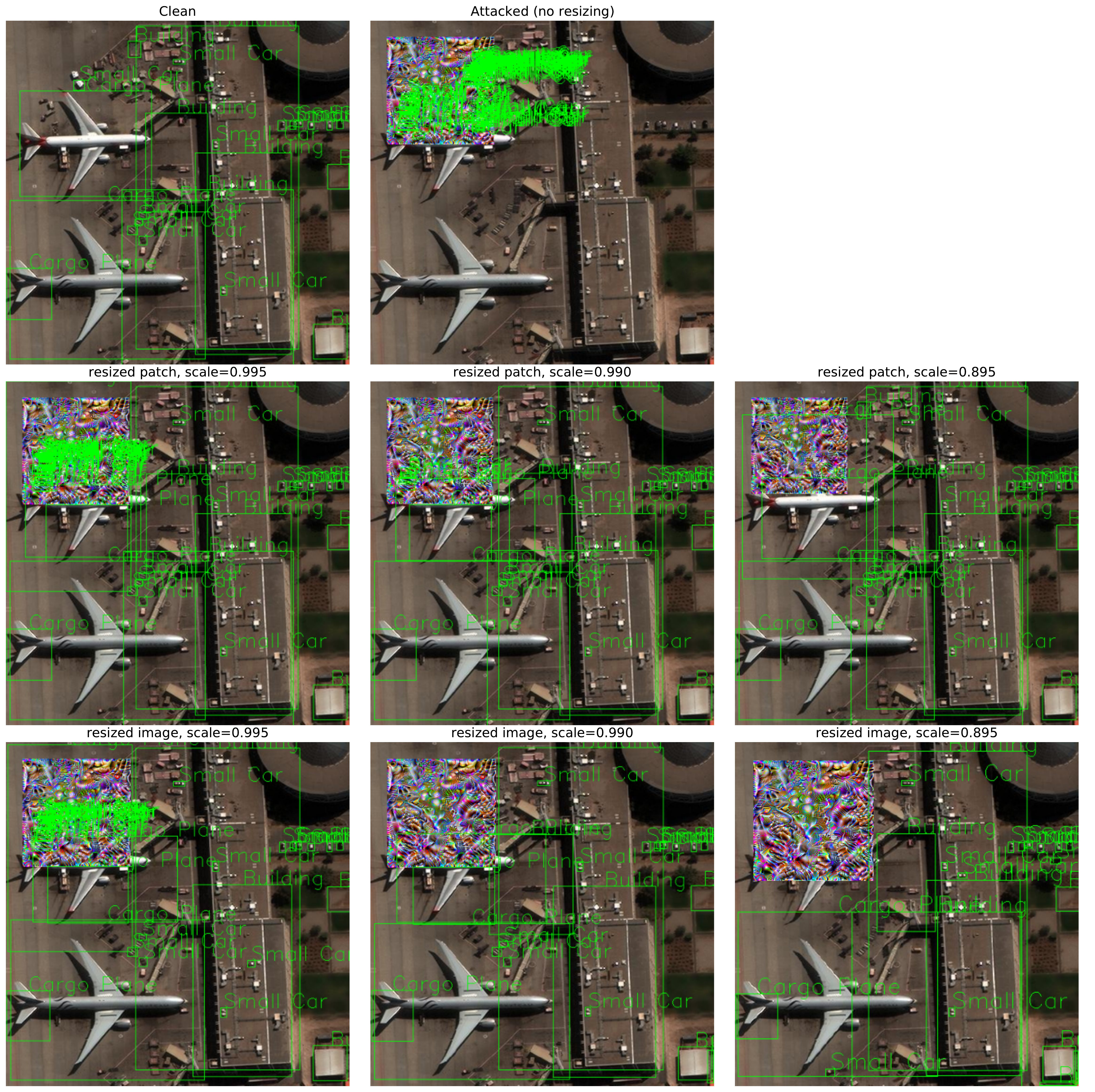}
    \caption{Effect of resizing. The first image corresponds to the predictions of the FRCNN that was fine-tuned on the xView data on a clean test set image, and the second image shows the predictions when an adversarial patch is digitally inserted into the image. The images in the second row correspond to resizing the patch while keeping the image fixed, and the images in the third row correspond to keeping the patch fixed and resizing the background image.}
    \label{fig:resize}
\end{figure}

In Fig.~\ref{fig:resize} we illustrate the effect of resizing both the patch and the background image.\footnote{A variety of resizing methods from the PILLOW library were considered, including NEAREST, LANCZOS, BILINEAR, BICUBIC, BOX, and HAMMING. These results correspond to the BOX method, as it caused the least amount of performance degradation.} The first image shows a clean test image, together with the predictions of the xXiew fine-tuned FRCNN. The second image shows the predictions when an adversarial patch is digitally inserted into the image. The images in the second row correspond to resizing the patch while keeping the image fixed, and the images in the third row correspond to keeping the patch fixed and resizing the background image. The figure shows that even for a very mild scale factor of 89.5 percent, the effect of the attack is completely ameliorated by resizing either the image or the patch. This implies that the efficacy of the patch is very sensitive to issues of scale. This finding motivated us to incorporate average scale transformations in the inner optimization loop of some of the patches, but this did not improve the results. Moreover, this experiment involved purely digital attacks, and it seems reasonable that the scale factors involved in the physical attack experiments described in the previous chapter are more extreme than 89.5 percent. Scale thus seems to be a plausible explanation for the failure of adversarial attacks to transfer to the physical world in our experiments.

\section{Discussion}

\subsection{Limitations}
Although we believe that the conclusions and lessons learned from our study hold rather generally, there are several limitations that should be documented. Most importantly, we had limited computational and engineering resources to devote to this project. We used existing neural network architectures and datasets that were loosely relevant to the testing environment, and all the models were trained relatively cheaply, with each model requiring between 1-8 hours of GPU time. 
Our models could have been improved by several ways. For example, the models could have been trained from scratch, rather than fine-tuning pre-trained models, bespoke neural architectures could have been designed, and perhaps, most significantly, the models could have been improved by investing in the curation of much larger training datasets. Unfortunately, all of these improved approaches require significant resources and were well beyond the scope of this project.

There are two important comments to make regarding the limited resources of the current study. First, we suspect that even with vastly greater resources it would still be very difficult if not impossible to generate adversarial patches that worked under a wide range of real-world conditions. Of course, this is conjecture based on this experiment, and it would be interesting to test more robustly in future work. Second, even if the results do qualitatively change when better models are trained with more data, it is reasonable to expect that many practical implementations of aerial or satellite-based object detection models will be similarly constrained to use pre-trained models and modestly-sized fine-tuning datasets.

Other limitations include the fact that we only considered one type of adversarial attack (the Lee-Kolter attack) and we did not optimize over the many hyper-parameters and implementation choices involved in deploying the attacks in both the digital and physical worlds. For example, we did not explore different learning rate schedules or different patch sizes, and we only considered a narrow range of printing methods and poster materials. With one exception, we also did not implement an important aspect of the Lee-Kolter attack, the “Expectation Over Transformation” \cite{athalye2018synthesizing} technique wherein the inner loop of the patch optimization involves the application of randomly sampled image transformations. This technique was specifically designed to improve the real-world transferability of the attacks, so it is possible that incorporating this technique would have improved the attack efficacy. 
The one exception was the 160x160 xView patch, where we averaged over random scale transformations with a scale factor uniformly sampled from the interval [0.5, 2]. However, in this case we did not observe any noticeable improvement in the attack performance.

Throughout this report we have emphasized the point that in order for the threat posed by adversarial examples to be taken seriously, the efficacy of these ML-based attacks should exceed (or at least be comparable to) the efficacy of low-tech, non-ML-based attacks such as simple camouflage or decoys. Due to the limited scope of this project, we only investigated the former; in future work it would be good to conduct a more thorough comparison by testing both approaches. This issue is mitigated by our strong negative findings for physical patch attacks, which seem to have failed almost completely. However, as noted above, there are many other attack approaches and conditions we could have explored, and it is quite plausible at least some of these would be effective and worth comparing against traditional methods of deception. 

\subsection{Conclusion}
In this work we conducted an empirical evaluation of adversarial patch attacks against Faster RCNN object detection models. We specifically focused on overhead (aerial or satellite) imagery, and we considered both digital and physical attacks. Our main finding is a negative one: our physical patch attacks completely failed to induce errors, either false positives or false negatives, in the models during our experiment. It is difficult to prove a negative, and, to be clear, it is not our claim that no physical attack would work against any possible object detection model applied to overhead imagery. As detailed above, our experiment was subject to limitations, and it is quite possible that the results would change if these limitations were addressed. Rather, we conclude that adversarial attacks may be more difficult to operationalize than some past work would suggest, and therefore they do not seem to pose an immediate threat to safety-critical AI systems. 

In addition to testing the efficacy of the attacks, our experiment also allowed us to work through many practical considerations associated with implementing the attack, such as the physical size of the printed patch, the pixel resolution, the quality of paper, where the patch should be located and whether it should be fixed or mobile, etc. One insight this produced was the observation that, in addition to not working well (or at all), the adversarial patch attacks are difficult to implement. The difficulty of implementation is compounded by the existence of other, tried-and-true alternative approaches, such as simple camouflage. This discussion aligns well with the critiques of \cite{gilmer2018motivating}, who argue that many papers on the security implications of adversarial examples ignore the existence of simpler and more effective attacks - the canonical example being that a stop sign detector could perhaps be fooled by attaching an adversarial sticker, but it could also be fooled by simply covering the sign with a bag or even knocking it down. Research into AI and adversarial examples is progressing at a rapid rate, but for now at least, we find that adversarial examples are not a major security threat to the overhead imagery scenario we considered.

\section*{Appendix: FRCNN Fine-Tuning Procedure}
In this appendix we give additional technical details on the fine-tuning of the FRCNN object detection models. We began with a FRCNN model that was pre-trained on the Common Objects in Context (COCO) dataset \cite{lin2014microsoft}. Starting with a pre-trained model allowed this study to be carried out using only modest resources, both in terms of the compute needed to fine-tune models on more relevant datasets and in terms of the costs needed to acquire enough images to train a reasonably good model. As discussed in above, we separately fine-tuned the pre-trained model on three additional datasets (xView, COWC, and OATML). In each case we used the same fine-tuning procedure. First, the data was split into a training and a testing set. For the xView and COWC datasets, this was done by randomly partitioning the data into an 80/20 split. For the OATML dataset, we used the same procedure described in \cite{hartnett2020operationally}. Next, because the pre-training and fine-tuning datasets have different numbers of classes (COCO: 90, xView: 60: COWC: 4: OATML: 1), the pre-trained FRCNN model must be modified by replacing the final neural network layer responsible for classification, the “ROI Heads Box Predictor”. This single layer is replaced with a randomly initialized layer with appropriate dimension for the fine-tuning dataset.

\begin{figure}
    \centering
    \includegraphics[width=0.5\textwidth]{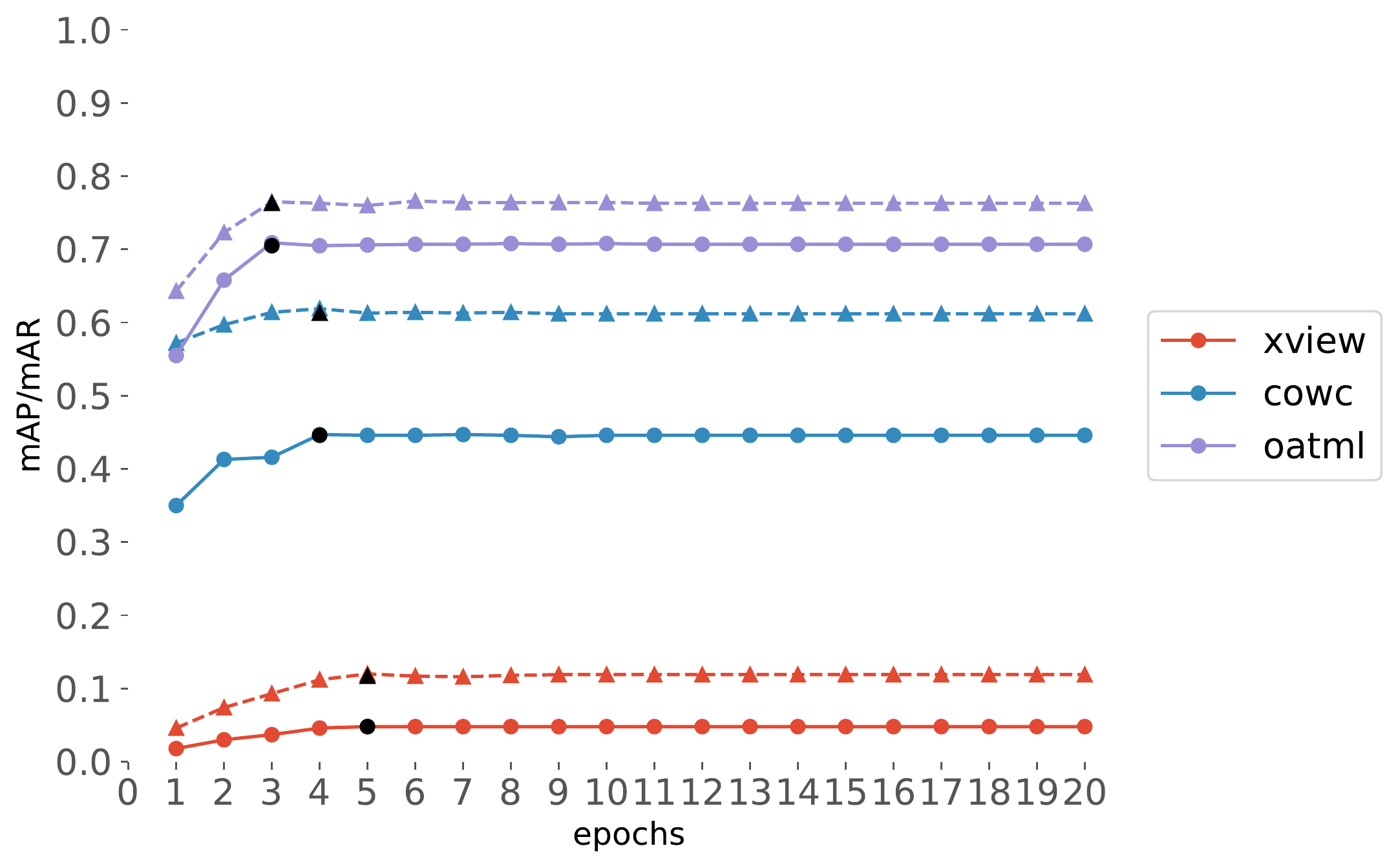}
    \caption{Test set performance of the FRCNN model evaluated after each fine-tuning training epoch. Curves are shown for each of the three training datasets, xView, COWC, and OATML. The mAP score is plotted with solid lines/closed circle markers and the mAR score is plotted with dashed lines and triangle markers. Among all the checkpoints, the best performing model according to the mAP score (shown in black) was selected to use with the adversarial attacks in the next section.}
    \label{fig:frcnnfinetuning} 
\end{figure}

There are two common fine-tuning approaches. In the first, only the weights of the new and randomly initialized layer are modified, and the other neural network parameters are kept unchanged. In the second, all the model parameters are simultaneous modified. We used the second approach. The training details are as follows. We used the built-in PyTorch implementation of stochastic gradient descent (SGD) with a learning rate of 0.002 , momentum set to 0.9, and a weight decay constant of 0.0005. We also used a learning rate scheduler which decreased the learning rate by a factor of 10 after 3 epochs. We used a fixed batch size of 6 and trained for 20 epochs for all models. It is important to note that holding the number epochs fixed implies that the total number of SGD updates varied between models, as each dataset contained a different number of images. At the end of each epoch the current model parameters were saved as checkpoints and the performance of the models, as measured in mean Average Precision (mAP) and mean Average Recall (mAR) were evaluated on the testing set.  These results are shown in Fig.~\ref{fig:frcnnfinetuning}. Curves are shown for each of the three training datasets, xView, COWC, and OATML. The mAP score is plotted with solid lines/closed circle markers and the mAR score is plotted with dashed lines and triangle markers. Among all the checkpoints, the best performing model according to the mAP score (shown in black) was selected to use with the adversarial attacks in the next section. 

\section*{Acknowledgments}
We acknowledge the support of Joel Predd, Director of the RAND National Security Research Division Acquisition and Technology Policy Center, and Michael Chaykowsky, who contributed to the early stages of this project. We also graciously acknowledge Marcus Hunt, the drone photographer. Lastly, we thank Christian Johnson and Justin Grana for providing helpful comments after reviewing an earlier version of the report.

\bibliography{refs}
\bibliographystyle{JHEP}

\end{document}